\def\year{2019}\relax
\documentclass[letterpaper]{article} %DO NOT CHANGE THIS
\usepackage{aaai19}  %Required
\usepackage{times}  %Required
\usepackage{helvet}  %Required
\usepackage{courier}  %Required
\usepackage{graphicx}  %Required
\usepackage{xcolor}  %Required
\usepackage{xspace}  %Required

\usepackage[hidelinks,breaklinks=true]{hyperref}
\usepackage{amsmath}
\usepackage{amssymb}
\usepackage{amsthm}
\usepackage{enumerate}
\usepackage{booktabs}
\newcommand{\Omit}[1]{}

\newcommand{\tup}[1]{\langle #1 \rangle}

\newcommand{\set}[1]{\ensuremath{\left\{#1 \right\}}}

\newcommand{\abs}[1]{\ensuremath{\left\vert{#1}\right\vert}}
\newcommand{\citeay}[1]{\citeauthor{#1} (\citeyear{#1})}

\newtheorem{definition}{Definition}

\newtheorem{theorem}[definition]{Theorem}

\newenvironment{example}{\noindent\textbf{Example.}\xspace}{\qed}

\newcommand{\Q}{\mathcal{Q}}
\newcommand{\F}{\mathcal{F}}
\renewcommand{\S}{\mathcal{S}}
\newcommand{\G}{\mathcal{G}}

\newcommand{\Eff}{{\mathit{Eff}}}

\newcommand{\abst}[2]{\tup{#1;#2}}
\newcommand{\Rule}[2]{\ensuremath{#1 \Rightarrow #2}}

\newcommand{\pplus}{\hspace{-.05em}\raisebox{.15ex}{\footnotesize$\uparrow$}}
\newcommand{\mminus}{\hspace{-.05em}\raisebox{.15ex}{\footnotesize$\downarrow$}}

\providecommand{\noopsort}[1]{}

\title{Learning Features and Abstract Actions for Computing Generalized Plans}

\author{%
Blai Bonet\\ Universidad Sim\'on Bol\'{\i}var \\ Caracas, Venezuela \\ bonet@usb.ve
\And Guillem Franc\`{e}s \\ University of Basel \\ Basel, Switzerland \\ guillem.frances@unibas.ch
\And Hector Geffner \\ ICREA \& Universitat Pompeu Fabra \\ Barcelona, Spain \\ hector.geffner@upf.edu
}

\begin{document}

\maketitle

\begin{abstract}
  Generalized planning is concerned with the computation of plans that solve not one but  multiple instances
  of a planning domain. Recently, it has been shown that generalized plans can be expressed as mappings of feature
  values into actions, and that they can often be computed with fully observable non-deterministic
  (FOND) planners. %  on suitable FOND problems. %after a sequence of transformations.
  The actions in such plans, however, are not the actions in the instances themselves, which are
  not necessarily common to  other instances, but \emph{abstract} actions that are defined on a set
  of \emph{common} features. The formulation assumes that the features and the abstract
  actions are given.
  In this work, we address this limitation by showing how to learn them automatically.
  %  the features and  abstract actions.
  The resulting account of generalized planning combines learning and planning
  in a novel way: a \emph{learner}, based on a Max SAT formulation, yields the features and abstract
  actions from sampled state transitions, and a FOND \emph{planner}
  uses this information, suitably transformed, to produce the general plans.
  Correctness guarantees are given and experimental results on several domains are reported.
\end{abstract}

\section{Introduction}

Generalized planning studies the computation of plans  that   solve   multiple  instances
%\cite{srivastava08learning,bonet09automatic,srivastava:generalized,hu:generalized,BelleL16,anders:generalized}.
\cite{srivastava08learning,bonet:icaps2009,hu:generalized,BelleL16,anders:generalized}.
For example, the plan  that iteratively  picks  a  clear block above $x$
and places it  on the table, achieves the goal $clear(x)$
in \emph{any} instance of the Blocksworld where the gripper is initially empty.
Once this  general plan or  policy is derived, it can be  applied to solve an infinite collection
of instances that involve different initial states, different objects, and  different (ground)  actions.

In the basic formulation due  to \citeay{hu:generalized}, a generalized plan
is a mapping of observations into actions that are assumed to be common
among  all the instances. More recently, this formulation has been extended by
\citeay{bonet:ijcai2018} to account for relational domains like Blocksworld  where the sets of
objects and  actions change from instance to instance.
In the new formulation, the observations  are replaced by  a set of boolean
and numerical features $F$ and a set of \emph{abstract actions} $A_F$.
These abstract actions are  \emph{sound} and \emph{complete} if they track the
effects  of the actions on  the features in a suitable way.
The resulting generalized plans  map feature values into abstract actions, and soundness
ensures that the application of an abstract action can be mapped back into the application of
a concrete action with the same effect  over the features. Moreover, the form of the abstract actions
ensures that generalized plans can be computed using fully observable non-deterministic (FOND) planners,
once the generalized planning problem is transformed into a FOND problem.
%after some transformations.

\Omit{
Abstract actions can increase or decrease numerical variables  $n$
associated with the numerical features (e.g., number of blocks above $x$), and while
increments  are transformed into deterministic propositional  effects $n > 0$ where $n > 0$ is
the negation of the proposition $n=0$, decrements  are  transformed  into
non-deterministic (disjunctive)  propositional effects $n > 0 \, | \, n=0$.
}

Bonet and Geffner's formulation of generalized planning  assumes that the
features and abstract actions are given. In this work, we address this limitation
by  \emph{showing  how the features and abstract actions  can be learned from
the primitive predicates used to define the instances and from sampled state transitions.}
For example,  the general policy for achieving $clear(x)$ % of \citeay{bonet:ijcai2018}
is obtained using a FOND planner on an abstraction that consists of  a  boolean
feature $H$,  that tracks whether the gripper holds a block,  a numerical feature
$n(x)$ that counts the number of blocks above $x$, and two abstract actions:
one with preconditions $\neg H$ and $n(x) > 0$, and effects $H$ and $n(x)\mminus$
(decrement of $n(x)$), and the other with precondition $H$ and effect $\neg H$.
We here show how to obtain such policies from STRIPS instances alone,
without having to provide the features and the abstract actions by hand.

This work relates to a number of research threads in planning, knowledge
representation, and machine learning. We make use of SAT solvers and description
logics for learning features and abstract actions. %\cite{sat-handbook,dl-handbook}.
The abstract actions provide a model from which the plans are obtained via
transformations and FOND planners \cite{geffner:book,ghallab:book}.
The model is not the model of an instance but a generalized model for obtaining
plans that work for multiple instances. In this sense, the work is different than
action model learning \cite{yang:action-learning} and model-based reinforcement
learning and closer in aims to work on learning general policies from examples or
experience \cite{martin-geffner:generalized,fern:bias,mazebase,general-drl}.
Generalized planning has also been formulated as a problem in first-order logic
\cite{srivastava:generalized}, and general plans over finite horizons have been
derived using first-order regression \cite{boutilier2001symbolic,wang2008first,van2012solving,sanner:practicalMDPs}.
Our approach differs from first-order approaches in the use of propositional
planners, and from purely learning approaches in the formal guarantees that
are characteristic of planning: if the learned abstract actions are  sound,
the resulting general plans must be correct.

The paper is organized as follows. We first provide the relevant background, and then show
how features and abstract actions can be learned by enforcing soundness and
completeness over a set of samples, from a pool of candidate features derived
from the domain predicates. The computational model is then summarized,
followed by experimental results and a discussion.

\section{Background}

We review generalized planning, abstract actions, and solutions following  \citeay{bonet:ijcai2018}.

\subsection{Generalized Planning}

A \emph{generalized planning problem} $\Q$ is a collection of planning instances $P$.
An instance $P$ is a  classical planning problem expressed in some compact language
as a tuple $P=\tup{V,I,G,A}$ where $V$ is a set of state variables that can take a
finite set of values (boolean or not), $I$ is a set of atoms over $V$ defining an
initial state $s_0$, $G$ is a set of literals over $V$ describing the goal
states, and $A$ is a set of actions $a$ with their preconditions and effects, which
define the set $A(s)$ of actions applicable in any state $s$, and the successor
state function $f(a,s)$, for any state $s$ and $a \in A(s)$.
A state is a valuation over $V$, and a solution to $P$ is an applicable action
sequence $\pi=a_0,\ldots,a_n$ that generates  a state sequence $s_0,s_1,\ldots,s_{n}$
where $s_n$ is a goal state (makes $G$ true). In this sequence, $a_i \in A(s_i)$
and $s_{i+1}=f(a_i,s_i)$ for $i=0, \ldots, n-1$. A state $s$ is reachable in $P$
if $s=s_n$ for one such sequence. A solution to the \emph{generalized problem}
$\Q$ is a solution to all instances $P \in Q$. The form of such solutions is
described below.

A \emph{feature} $f$ for a class $\Q$ of problems represents a function $\phi_f$
that takes an instance $P$ from $\Q$ and a state $s$ reachable in $P$, and results
in a value $\phi_f(s)$.
A feature is \emph{boolean} if it results in boolean values, and \emph{numeric}
if it results in numerical values, here assumed to be non-negative.
For example, $H$ and $n(x)$ are two features in Blocksworld:
$H$, boolean, tracks whether the gripper is empty, while
$n(x)$, numerical, tracks the number of blocks above $x$.

Symbols like $x$ denote \emph{parameters} whose value depends on the
instance $P$ in $\Q$. For example, if $\Q_{clear}$ denotes the Blocksworld
instances with goals of the form $clear(x)$, then a problem
$P$ with goal $clear(A)$ will belong to $\Q_{clear}$ and
the value of $x$ in $P$ will be (the block) $A$.

While instances $P$ in a generalized problem $\Q$ would normally share some structure \cite{bonet:ijcai2015},
like the same planning domain and predicate symbols, this is not strictly  necessary.
On the other hand,  the features $f$  must be common to all the instances  in $\Q$ and
represent functions $\phi_f(s)$ that are well defined over all reachable states $s$.

\subsection{Abstract Actions}

An \emph{abstract action}  for a generalized problem $\Q$ and a set $F$ of features
is a pair $\bar{a}=\abst{Pre}{\Eff}$ where $Pre$ and $\Eff$ are the  preconditions
and effects expressed in terms of the  set $V_F$ of boolean and numerical \emph{state variables}
associated with the features. Preconditions and effects over boolean state variables $p$ are literals of the form $p$
and $\neg p$, that abbreviate the  atoms $p=true$ and $p=false$, while
preconditions and effects over  numerical state variables $n$ are of  the form $n=0$ and $n > 0$,
and   $n\mminus$ (decrements) and  $n\pplus$ (increments), respectively.
The language for the abstraction,  that combines boolean and numerical variables
that can be decreased or increased by unspecified amounts, is the language of \emph{qualitative numerical problems (QNPs)}.
Unlike standard numerical planning problems \cite{helmert:numerical}, QNPs
are decidable and can be solved effectively by means of  FOND planners \cite{srivastava:aaai2011,bonet:ijcai2017}.

Features $f$ refer to \emph{state functions} $\phi_f(s)$ in the instances $P$
of the generalized problem $\Q$, but  to \emph{state variables} in  the abstraction.
An \emph{abstract state} is  a \emph{truth valuation} over
the atoms $p$ and $n=0$ defined over the state variables $p$ and $n$
associated with the boolean and  numerical features.
The abstract state $\bar{s}$ that \emph{corresponds} to a concrete state $s$
in an instance $P$ of $\Q$ is the truth valuation that makes $p$ true iff $\phi_p(s)=true$,
and $n=0$ true iff $\phi_n(s) = 0$. An abstract action $\bar{a}$ is applicable in  $s$
if its preconditions are true in $\bar{s}$.

An action $a$ and an abstract action $\bar{a}=\abst{Pre}{\Eff}$  \emph{have  the same qualitative effects over the features $F$
in a state $s$} when both are applicable in $s$ with the same effects on the boolean features and
the same \emph{qualitative effects} on the numerical features. If $s'$ is the result of applying $a$ in $s$,
this means that \cite{bonet:ijcai2018}:
1)~$p \in \Eff$ and $\neg p \in Pre$ iff
$\phi_p(s)$ is false and $\phi_p(s')$ is true (i.e., $p$ becomes true),
2)~$\neg p \in \Eff$ and $p \in Pre$ iff
$\phi_p(s)$ is true and $\phi_p(s')$ is false ($p$ becomes false),
3)~$n\pplus \in \Eff$ iff $\phi_n(s') > \phi_n(s)$ ($n$ increases),
and 
4)~$n\mminus \in \Eff$ iff $\phi_n(s') < \phi_n(s)$ ($n$ decreases).

\medskip
\begin{example}
Let $\Q_{clear}$ stand for  all  Blocksworld instances with stack and unstack actions and goal $clear(x)$,
let $F=\{H,n(x)\}$ be the set with the two features above, and let $s$ be a reachable state in an instance $P$
in $\Q_{clear}$
where the gripper is empty, the atoms $on(A,B)$ and $clear(A)$ are true, and the block $A$ is above $x$.
In this state $s$,
the abstract action $\bar{a}=\abst{\neg H,n(x)>0}{H,n(x)\mminus}$
and the concrete  action  $a=\text{Unstack}(A,B)$ have the same effect over the features.
Indeed, $\bar{a}$ and $a$  are both applicable in $s$, and while $\bar{a}$ makes the variable $H$ true and decreases
$n(x)$, the action $a$ results in a state $s'$ where $\phi_{H}(s')$ is true and $\phi_{n(x)}(s') < \phi_{n(x)}(s)$.
\end{example}

\subsection{Sound and Complete Abstractions}

Soundness and completeness are the key  properties that  enable us to reason about the collection of
instances $\Q$ in terms of abstract actions that operate at the level of the features:

\begin{definition}
A set of abstract actions $A_F$ over the features $F$ is \emph{sound} relative to
$\Q$ iff for any reachable state $s$ over an   instance $P$ in $\Q$,
if an abstract action $\bar{a}$ in $A_F$ is applicable in $\bar{s}$,
there is an  action $a$ in $P$ that is applicable in $s$ and
has the \emph{same qualitative effects} over the features as $\bar{a}$.
\end{definition}

\begin{definition}
A set of abstract actions $A_F$ is \emph{complete} iff for any
reachable state $s$ over an instance $P$ in $\Q$ and any  action $a$
in $P$ that is applicable in $s$, there is an abstract action $\bar{a}$ in
$A_F$ that is applicable in $\bar{s}$ and  has the \emph{same qualitative
effects over the features as $a$.}
\end{definition}

Let us say that  action $a$ \emph{instantiates} abstract action $\bar{a}$ in a state $s$,
and $\bar{a}$ \emph{captures} $a$ in $s$,  when $a$ and $\bar{a}$ are both applicable in $s$ and have the same qualitative effects over the features.
Then, soundness means that any abstract action that is applicable in some reachable state can
always be instantiated by a concrete action, while completeness  means that any concrete action that is applicable in some reachable state
is always captured by an abstract action.

It may well be possible  that a  feature set $F$ does not  support a set of abstract actions $A_F$
that is both sound and complete relative to $\Q$. Soundness, however, is  crucial for deriving general plans that are valid.
An important intuition is that the features in $F$ support a set of sound abstract actions $A_F$ when the
effects of the concrete actions over $F$ can be \emph{predicted};
that is, the qualitative values of the features in a state $s$ of an instance $P$,
as captured by $\bar{s}$, \emph{determine} the possible ways in which the actual
values of the features may change in the transitions from $s$ to a successor state $s'$.
Hence, if $a$ and $b$ are two actions applicable in $s$ such that one makes
$p$ true and the other increases the value of $n$, then in any other state $t$
such that $\bar{t}=\bar{s}$, there should be applicable actions $a'$ and $b'$
with the same qualitative effects on $p$ and $n$.

\medskip
\begin{example}
  The abstract action set $A_F = \{\bar{a}, \bar{a}'\}$ where $\bar{a} = \abst{\neg H,n(x)>0}{H,n(x)\mminus}$
  and $\bar{a}'=\abst{H}{\neg H}$, defined over the feature set $F=\{H,n(x)\}$,
  is sound but not complete relative to $\Q_{clear}$.
  The actions in $A_F$ capture the concrete actions that pick blocks from above $x$ and put them away,
  but not other actions like stacking a block above $x$, or picking a block that is not above $x$.
\end{example}

\subsection{Solutions}

A solution to a problem $\Q$ given the features $F$ and abstract
actions $A_F$ is a partial function $\pi$ that maps \emph{abstract states}
into \emph{abstract actions} such that $\pi$ solves all instances $P$ in $\Q$.
The plan or policy $\pi$ induces a trajectory $s_0,a_0,s_1, \ldots, s_n$ in
an instance $P$ of $\Q$ iff
1)~$s_0$ is the initial state in $P$,
2)~$a_i$ is one of the actions that instantiate $\pi(\bar{s}_i)$ in $s_i$,
3)~$a_i$ is applicable in $s_i$ and $s_{i+1}=f(a_i,s_i)$, and
4)~$s_n$ is a goal state of $P$, or $\pi(\bar{s}_n)$ is undefined,
not applicable in $\bar s_n$, or no applicable action $a_n$ in $P$ instantiates $\pi(\bar{s}_n)$.
The policy $\pi$ solves $P$ iff all trajectories induced by $\pi$ reach a goal state of $P$.

\medskip
\begin{example}
  A solution for $\Q_{clear}$ with   $F=\{H,n(x)\}$ is the policy $\pi$ given by the rules
  $\Rule{\neg H, n(x)>0}{\bar{a}}$ and $\Rule{H, n(x)>0}{\bar{a}'}$
  for the abstract actions $\bar{a}$ and $\bar{a}'$ above.
  The policy picks blocks above $x$ and puts them aside (never  above $x$)
  until $n(x)$ becomes zero.
\end{example}

\subsection{Computation}

The steps for obtaining policies $\pi$ for a generalized problem $\Q$
using the features $F$ and abstract actions $A_F$ are as follows \cite{bonet:ijcai2018}:

\begin{enumerate}[1.]
  \item The state variables $V_F$ and the abstract actions $A_F$ are
    extended with initial and goal formulas $I_F$ and $G_F$ over $V_F$
    to yield the \emph{abstraction}  $Q_F=\tup{V_F,I_F,G_F,A_F}$,
    which is a QNP. For soundness, $I_F$ must be such that
    the initial states of instances $P$ in $\Q$ all satisfy $I_F$, while
    all states that satisfy $G_F$ must be goal states of $P$.
    %\textcolor{red}{\bf ** What is the format for $I_F$ and $G_F$? Conjunctions of lits, DNF, ...? **}
%
  \item The abstraction $Q_F=\tup{V_F,I_F,G_F,A_F}$ is converted into a
    \emph{boolean FOND} problem $Q'_F=\tup{V'_F,I'_F,G'_F,A'_F}$ by
    replacing the numerical variables $n\in N$ by the symbols $n=0$,
    the first-order literals $n=0$ by propositional literals  $n=0$,
    the effects $n\pplus$ by effects $n > 0$, and the
    effects $n\mminus$ by non-deterministic effects $n>0\,|\,n=0$.
  \item The solutions computed by a FOND planner on $Q'_F$ are the
    strong cyclic solutions of $Q'_F$ \cite{strong-cyclic}.
    Such solutions however do not necessarily solve $Q_F$ because the
    non-deterministic effects $n>0\,|\,n=0$ in $Q'_F$ are not \emph{fair} but
    \emph{conditionally fair}: infinite decrements of $n$ ensure that
    $n=0$ is true eventually but \emph{only} when  the number of increments of $n$ is finite.
    The problem of obtaining solutions of $Q'_F$ that only assume conditional
    fairness, the so-called terminating solutions \cite{srivastava:aaai2011},
    is mapped into the problem of solving an amended FOND $Q^+_F$ that
    assumes standard  fairness \cite{bonet:ijcai2017}.\footnote{Translator
    available at https://github.com/bonetblai/qnp2fond.}
    % translation to $Q^+_F$ is performed with software supplied by the authors.}
    %available by the authors that fixes bugs in the description found]
       %in the paper at https://github.com/bonetblai/qnp2fond.}
\end{enumerate}

\begin{theorem}[\citeay{bonet:ijcai2018}]
If the abstract actions  $A_F$ are sound relative to the generalized problem $\Q$, the  solutions  to $Q^+_F$
computed by FOND planners off-the-shelf  are solutions to $\Q$.
\end{theorem}

\medskip
\begin{example}
  Let us restrict  $\Q_{clear}$  to those  instances where  the gripper is initially empty
  and there are blocks on top of $x$. For $F=\{H,n(x)\}$  and $A_F=\{\bar{a},\bar{a}'\}$ as above,
  $Q_F=\tup{V_F,I_F,G_F,A_F}$ may be defined with $I_F=\{\neg H, n(x) > 0\}$ and $G_F=\{n(x)=0\}$.
  $Q'_F$ is like $Q_F$ but with $n(x)=0$ regarded as a propositional symbol, $n(x) > 0$
  as its negation, and the effect $n(x)\mminus$ replaced by  $n(x) > 0 \, | \, n(x)=0$.
  Since  no action in $A_F$ increases $n(x)$, the strong solutions of $Q'_F$
  are solutions to $\Q_{clear}$ \cite{bonet:ijcai2017}.
  %$Q^+_F=Q'_F$, and the strong cyclic solution
  %of $Q'_F$, that corresponds to the policy above, solves $\Q_{clear}$.
  %\textcolor{red}{\bf ** Why do we need to enter into details about $Q^+_F$?
  %We are using a tool and should not worry about such details.
  %Otherwise, we are conveying the idea that the user should worry about them and
  %then know how the QNP2FOND translation work. **}
\end{example}

\section{Approximate Soundness and Completeness}

The formulation and computational model above, from \cite{bonet:ijcai2018},
assume that the features $F$ and abstract actions $A_F$ are given.
The contribution of this work is a \emph{method for  learning   them automatically
by enforcing a form of soundness and completeness over a set of samples:}

\begin{definition}
  For a generalized problem $\Q$, a \emph{sample set}  $\S$ is a non-empty set of tuples  $(s,s',P)$
  such that 1)~$P$ is an instance of $\Q$, 2)~$s$ is a reachable state in $P$, 3)~$s'$
  is a successor of $s$ in $P$; i.e., $s'=f(a,s)$ for some action $a \in A(s)$,
  and
  4)~$\S$ is closed in the sense  that if $(s,s',P) \in \S$, then $(s,s'',P) \in \S$  for any successor $s''$
  of $s$ in $P$.
\end{definition}

By assuming that sampled states $s$ are tagged with the instance $P$,
we abbreviate the tuples $(s,s',P)$ as $(s,s')$, and refer to the
states $s'$ in  the pairs $(s,s') \in \S$ as the successors of
$s$ in $\S$. The  closure condition 4)~requires that
states $s$ appearing \emph{first} in pairs $(s,s')$ must be fully expanded in $\S$;
namely, all possible transitions $(s,s'')$ must be  in the sample. We call them
the \emph{expanded} states in $\S$.

For defining soundness and completeness \emph{over a sample set}, we say that  a transition $(s,s')$ and an abstract action $\bar{a}$
have  the same qualitative effects over the features,   when in the  state $s$,  the action $a$ that maps $s$ into $s'$ and the abstract action
$\bar{a}$ have the same qualitative effects over the features:

\begin{definition}
  A set of abstract actions $A_F$ is sound \emph{relative to a  sample set} $\S$ for $\Q$
  iff for  any  abstract action $\bar{a}$ in $A_F$ applicable in an expanded state $s$ of $\S$,
  there is a transition $(s,s') \in \S$ with the same qualitative  effects over $F$ as $\bar{a}$.
\end{definition}

\begin{definition}
  A set of abstract actions $A_F$ is complete   \emph{relative to a sample set} $\S$ for $\Q$
  iff for each transition  $(s,s')$ in  $\S$, there is an abstract action $\bar{a}$ in $A_F$
  with the same qualitative effects over $F$ that is applicable in $s$.
\end{definition}

For a sufficiently large sample set, the approximate and exact notions of soundness and completeness   converge.

\section{Learning Features and Abstract Actions}

In order to learn the features and abstract actions from a sample set $\S$ for $\Q$,
we define a propositional formula $T(\S,\F)$, where  $\cal F$ represents a large
pool of candidate features, that is \emph{satisfiable} iff there is
a set of abstract actions $A_F$ over  $F \subseteq \F$ that is sound and complete
relative to $\S$.

\subsection{SAT Encoding: $T(\S,\F)$ and $T_G(\S,\F)$}

For each transition  $(s, s') \in \S$ and each  feature $f \in \F$,
$\Delta_f(s, s') \in \{+, -, \uparrow, \downarrow, \bot\}$  denotes the
\emph{qualitative change of value} of feature $f$ along the transition $(s, s')$,
which can go from false to true ($+$), from true to false ($-$) or remain
unchanged ($\bot$), for boolean features, and can increase ($\uparrow$),
decrease ($\downarrow$), or remain unchanged ($\bot$), for numerical features.
The \emph{propositional variables} in $T(\S,\F)$ are then:
\begin{enumerate}[{\small$\bullet$}]
  \item $selected(f)$ for each $f \in \F$, true iff $f$ selected (in $F$).
  \item $D_1(s,t)$ for states $s$ and $t$ expanded in $\S$,  true iff the
    selected features distinguish $s$ from $t$; i.e., if $s$ and $t$ disagree
    on the truth value of some selected feature, either a boolean or numeric feature,
    %on  the truth value of atom  $p$ or  $n=0$ for   selected  feature $p$ or $n$.
    %Since $D_1$ is symmetric, only symbols for $s < t$ are created for some
    %arbitrary, fixed order over the states in $\S_1$.
  \item $D_2(s, s', t, t')$ for each $(s, s')$ and $(t, t')$ in $\S$,
    true iff some selected feature $f$ distinguishes the two transitions; i.e.,
    $\Delta_f(s, s')\not=\Delta_f(t,t')$ for some selected feature $f$.
    %As for $D_1$, the number of $D_2$ variables is cut in half due to symmetries.
\end{enumerate}

\medskip
\noindent The first formulas in $T(\S,\F)$ capture the meaning of $D_1$:
\begin{equation}
  \label{eq:d1}
  D_1(s, t) \ \Leftrightarrow\ \textstyle \bigvee_{f}  selected(f)
\end{equation}

\noindent where $f$ ranges over the features in $\F$ with  different qualitative
values in $s$ and $t$; namely, boolean features $p$ with different
truth values in $s$ and $t$, and numerical features $n$ for which
the atom $n=0$ has different truth values in $s$ and $t$.

The second class of formulas encode the meaning of $D_2$:
\begin{equation}
  \label{eq:d2}
  D_2(s, s', t, t') \ \Leftrightarrow\ \textstyle\bigvee_f  selected(f)
\end{equation}

\noindent where $f$ ranges over the features in $\F$ that have the same
qualitative values in $s$ and $t$ but which change differently in the two
transitions; i.e., $\Delta_f(s, s') \neq \Delta_f(t, t')$.

The third class of formulas relate $D_1$ and $D_2$ by enforcing \emph{soundness}
and \emph{completeness} over  the sample: %, expressed as
\Omit{
Since the abstract actions  are not given,
the qualitative changes in each  transition $(s,s')$ in $\S$
are  taken as templates  of abstract actions. Thus, if $s$ and $t$ are not distinguished by the selected features,
for each transition $(s, s') \in \S$, there must be a transition $(t, t') \in \S$
such that the two transitions are not distinguished by the selected features either.
This is expressed as
}
\begin{equation}
  \label{eq:bridge1}
  \neg D_1(s, t) \  \Rightarrow\ \textstyle\bigvee_{t'} \neg D_2(s, s', t, t')
\end{equation}

\noindent where $s$ and $t$ are expanded states in  $\S$, $s'$ is a successor of $s$ in $\S$,
and $t'$ ranges over the successors of $t$ in $\S$. This formula is crucial.
It says that \emph{if the selected features do not distinguish state $s$ from $t$,
then for each action $a$ that maps  $s$ into $s'$, there must be an action $b$ that maps the state
$t$ into a state $t'$ such that the two transitions $(s,s')$ and $(t,t')$ affect the selected features
in the same way.} The formula does not mention the actions $a$ and $b$ because their identity does not matter;
it is just the state transitions that count. In addition, the formula does not involve abstract actions as they
will be obtained from the transitions and  the satisfying assignment. Indeed, for each transition $(s,s')$ in the sample, there will be
an abstract action $\bar{a}$ that  accounts for the transition; i.e., with the  same qualitative effects over the selected features.
Moreover, if $D_1(s,t)$ and $D_2(s,s',t,t')$ are both false in the satisfying assignment, the two transitions $(s,s')$ and $(t,t')$
will be captured by the same abstract action.

The fourth and  last class of formulas force the selected features to distinguish goal from non-goal states as:
\begin{equation}
  \label{eq:goal}
  D_1(s,t)
\end{equation}

\noindent where  $s$ and $t$  are expanded states  in $\S$ such that exactly one of them is a goal state.
% For this, it is assumed that expanded states $s$ in $\S$ are labeled as goal or non-goal states.
%For this, it is assumed that states $s$ sampled from an instance $P$, are labeled as goal or non-goal states of $P$.
For this, it is assumed that the states in the sample are labeled as goal or non-goal states.

\medskip

The SAT theory $T(\S,\F)$ given by formulas \eqref{eq:d1}--\eqref{eq:goal}
has  $|\F| + m^2 (b^2 + 1)$ propositional variables,
%has a number of propositional variables equal to $m^2 \times (b^2 + 1) + |\F|$
where $m$ is the number of expanded states in $\S$, and $b$ is their average branching factor (transitions per state).
The number of clauses is bounded by $m^2(2 + b^2 + b + |\F|(b^2 + 1))$. %$m^2 \times b^2 \times |\F|$.
\Omit{
This means  that  $50$ expanded states in the sample set
with an average branching factor of $10$, and a set $\F$
with $400$ candidate features, can generate up to $10^8$ clauses.
While this is a loose upper bound for the number of clauses that
follow from the implication   $D_2(s, s', t, t') \Leftharrow\ \textstyle\bigvee_f  selected(f)$
in (\label{eq:d2}) and the numbers are further reduced by symmetry consideration,
this is still a large number. ..
}

We also consider an alternative SAT theory $T_G(\S,\F)$ that is similar to $T(\S,\F)$
but  smaller. It is obtained by marking  some transitions $(s,s')$ in $\S$  as  \emph{goal relevant}.
Then, rather than creating abstract  actions to account for all the transitions, we only create
abstract actions to account for the marked transitions. This is achieved by drawing the states
$s$ and the transitions $(s,s')$ in formulas \eqref{eq:d1} and \eqref{eq:d2} from the
set of marked transitions in $\S$. The states $t$ and the transitions $(t,t')$, on the other hand,
are drawn from the whole sample set $\S$ as before. This simplification preserves soundness over $\S$
but completeness is not over $\S$ but over the marked transitions in $\S$.
The goal-relevant transitions in $\S$ are obtained by computing one plan
for each sampled instance $P$ from $\Q$.
%The transitions marked as goal relevant in $\S$  are obtained by computing one plan
%for each sampled instance $P$ from $\Q$.

\Omit{
If the set of marked transitions is much smaller than the full sample set, %as will typically be the case,
$T_G(\S,\F)$ is much smaller than $T(\S,\F)$.
$T_G(\S,\F)$ guarantees soundness relative to $\S$
but completeness only relative to the subset of goal-relevant transitions in $\S$.
Yet this does not affect the resulting formal guarantees, which depend only on soundness.
}

\subsection{Extracting $F$ and  $A_F$}

For a satisfying assignment $\sigma$ of the theories $T(\S,\F)$ and $T_G(\S,\F)$, let
$F_{\sigma}$ be the set of features $f$ in $\F$ such that $selected(f)$
is true in $\sigma$, and let $A_{\sigma}$ be the set of abstract actions
that \emph{capture} all transitions $(s,s')$ in $\S$, in the case of theory $T$,
and the goal-relevant transitions $(s,s')$ in $\S$, in the case of $T_G$.
% with duplicates and redundancies removed.
The abstract action $\bar{a}=\abst{Pre}{\Eff}$ that captures the transition $(s,s')$
has the \emph{precondition} $p$ (resp.\ $\neg p$) if $p$ is a boolean feature in
$F_{\sigma}$ that is true (resp.\ false) in $s$, and has the precondition $n=0$
(resp.\ $n > 0$) if $n$ is a numerical feature in $F_{\sigma}$ such that $n=0$
is true (resp.\ false) in $s$. Similarly, $\bar{a}$ has the \emph{effect} $p$
(resp.\ $\neg p$) if $p$ is a boolean feature in $F_{\sigma}$ that is true
(resp.\ false) in $s'$ but false (resp.\ true) in $s$, and the effect $n\mminus$
(resp.\ $n\pplus$) if $n$ is a numerical feature in $F_{\sigma}$ whose values
decreases (resp.\ increases) in the transition from $s$ to $s'$.
Duplicate abstract actions are removed, and if two abstract actions $\bar{a}$ and $\bar{a}'$
differ only in the sign of a precondition, the two abstract actions are    merged into one with
the precondition dropped.
%The key properties of $T(\S,\F)$ and $T_G(\S,\F)$ are:

\begin{theorem}
  The theory $T(\S,\F)$ is satisfiable iff there is a set of features $F \subseteq \F$
  and a set of abstract actions $A_F$ over $F$ such that $A_F$ is sound and complete relative to  $\S$.
\end{theorem}

\begin{theorem}
  If $\sigma$ is a satisfying assignment of $T(\S,\F)$, the set $A_{\sigma}$ of abstract actions over $F_{\sigma}$ is sound and complete relative to $\S$.
\end{theorem}

\begin{theorem}
  If $\sigma$ is a satisfying assignment of $T_G(\S,\F)$, the set $A_{\sigma}$ of abstract actions over $F_{\sigma}$ is sound relative to $\S$ and complete relative to the marked transitions in $\S$.
  %If $\sigma$ satisfies  $T_G(\S,\F)$, $A_{\sigma}$  is sound relative to $\S$ and complete relative to the marked transitions in $\S$.
\end{theorem}

While the initial and goal expressions $I_F$ and $G_F$ of the abstract problem can be learned
from the satisfying assignment $\sigma$ as well, for simplicity, in the examples below they are
provided by hand.\footnote{$G_F = G_{\sigma}$ may be defined as the DNF formula whose terms
  correspond to the abstract states over $F_{\sigma}$ that correspond to goal states in
  the sample.
  % For each such term, a new abstract action to reach a new goal literal
  % from such a term may be required.
  }
% \footnote{$G_F = G_{\sigma}$ can be defined as the DNF formula
% whose terms  correspond to the abstract states over the selected features in $F_{\sigma}$
% that are true in some  goal state of the sample  and false in all non-goal states.}

\Omit{
Due to  \eqref{eq:d1} and \eqref{eq:goal} that force the selected features in $F_{\sigma}$
to   distinguish goal  from non-goal states, we have that:

\begin{theorem}
For a satisfying assignment $\sigma$  of $T(\S,\F)$,
$s$ is expanded goal   state in $\S$  iff  $s$ satisfies  $G_F=G_{\sigma}$.
\end{theorem}
}

Different assignments $\sigma$ result in different feature sets $F_{\sigma}$
and abstract action sets $A_{\sigma}$. Simpler and more meaningful features are found
by minimizing a cost measure such as $|F_{\sigma}|$
or, more generally, $\sum_{f \in F_{\sigma}} cost(f)$.
To achieve this, we cast our problem
into a constrained optimization problem, more specifically, a Weighted Max-SAT problem,
where clauses resulting from the theories $T$ or $T_G$ are taken as \emph{hard} clauses,
and we add, for each feature $f \in \F$, a
\emph{soft} unit clause $\neg selected(f)$ with weight $cost(f)$, defined below.

\section{Feature Pool}

The feature pool $\F$ used in $T(\S,\F)$ is obtained
from the   predicates encoding  the instances in $\Q$
that are assumed to be common to all such instances.
%For this, these predicates are assumed to be common to
%all such instances, and to have arity no greater than 2.
From these \emph{primitive predicates} and some composition rules,
we define a large set of \emph{derived predicates $r$ of arity $1$}
whose denotation $r^s$ in a state $s$ over an instance $P$
refers to the set of constants (objects) $c$ in $P$ that
have the property $r$ in $s$; i.e., $r^s=\{c \, | \, r(c) \hbox{ is true in $s$}\}$.
Boolean and numerical features $p_r$ and $n_r$ can then be defined from  $r$,
denoting the functions $\phi_{p_r}(s)$ and $\phi_{n_r}(s)$: the first is $0$
if $|r^s|=0$ and else is $1$, the second represents the cardinality $|r^s|$ of $r$ in $s$.

Unary predicates $r(x)$ can be created by definitions of the form ``$r(x)$ iff $\exists y[q(x,y) \land t(y)]$''
where $q$ and $t$ are  primitive  predicates of arity $2$ and $1$ respectively.
We  use   instead the syntax and  grammar
of  description logics \cite{dl-handbook}.  In description logics, unary and binary predicates
are referred to as \emph{concepts} $C$  and \emph{roles} $R$. Description logics  have been used
for capturing general policies using purely learning methods \cite{martin-geffner:generalized,fern:bias}.

\subsection{Concepts}

The concepts and roles denoted as $C_p$ and $R_p$ represent the \emph{primitive predicates} of arity 1 and 2 respectively.
The grammar for generating new concepts and roles from them is:

\begin{enumerate}[{\small$\bullet$}]
  \item $C_A \leftarrow C_p, C_u, C_x$,  primitive, universal, nominals: $C_u$ denotes  universe,
   $C_x$ denotes $\set{x}$ for parameter $x$ if any,
  \item $C \leftarrow \neg C_A$, negation on primitive, universal, nominals,
  \item $C \leftarrow C \sqcap C'$, conjunctions,
  \item $C \leftarrow \exists R.C,\forall R.C$, first denotes $\{x:\!\exists y[R(x,y) \land C(y)]\}$, %  $x$'s  s.t.\  $R(x,y) \land C(y)$ for some  $y$;
    the second denotes $\{x:\forall y[R(x,y)\land C(y)]\}$, %,  $x$'s  s.t. $R(x,y) \rightarrow C(y)$ for all $y$,
  \item $C \leftarrow R=R'$, denotes $\{ x : \forall y[R(x,y)=R'(x,y)]\}$, %$x$'s  s.t. $R(x,y)$ iff $R'(x,y)$ for all $y$,
  %  set of $y$'s  for which $R(x,y)$ equal to set of $y$'s s.t. $R'(x,y)$.
  \item $R \leftarrow R_p, R_p^{-1}, R_p^+, [R_p^{-1}]^+$: primitive, inverse, and transitive closure of both.
%     \item $R \leftarrow R \circ R'$, role composition: $(x,y)$ in $R \circ R'$ if $(x,z)$ in $R$ and $(z,y)$ in $R'$.
%     \item $R \leftarrow R:C$, denotes pairs $(x,y)$ in $R$ s.t. $C(y)$.
 \end{enumerate}

The denotations $C^s$ and $R^s$ for concepts and roles follows
from the rules and the denotation of primitive concepts and roles.
For example, for the concept $C: \exists\, on^+ . C_x$ in Blocksworld,
$C^s$ denotes the set of blocks that are above $x$.
%The numerical feature $n_C$ associated with concept $C$ represents
%the number of objects in $C^s$. %$|C^s|$.

\subsection{Candidate Features}

The \emph{complexity} of a concept or role is the minimum number of grammar  rules needed to generate it.
The set of concepts and roles with complexity no greater than  $k$ is referred to as  $\G^k$.
When generating  $\G^k$,  redundant  concepts are incrementally pruned. A concept is deemed
redundant when its  denotation coincides with the denotation of a previously generated concept
over all  states in $\S$.  The set $\F=\F^k$, defined from $\G^k$, contains the following features:
\begin{enumerate}[{\small$\bullet$}]
  \item For each nullary primitive predicate $p$, a \emph{boolean}
    feature $b_p$ that is true in $s$ iff $p$ is true in $s$.
  %
  %\item For each concept $C$, a \emph{boolean} feature $b_C$, if $|C^s|\leq 1$
  \item For each concept $C$, a \emph{boolean} feature $b_C$, if $|C^s| \in \set{0,1}$
    for all sampled states $s$, and a \emph{numerical} feature $n_C$ otherwise.
    The value of $b_C$ in $s$ is true iff $|C^s| > 0$; the value of $n_C$ is $|C^s|$.
  \item Numerical features $\textit{dist}(C_1,R{:}C,C_2)$ that represent the
    smallest $n$ such that there are objects $x_1, \ldots, x_n$ satisfying
    $C_1^s(x_1)$, $C_2^s(x_{n})$, and $(R{:}C)^s(x_i,x_{i+1})$ for $i=1,\ldots,n$.
    The denotation $(R{:}C)^s$ contains all pairs $(x,y)$ in $R^s$ such that $y\in C^s$.
\end{enumerate}

% We also prune from this space of features all features whose denotation is either constant
% or coincident with that of another feature over all states in $\S$.
The measure $cost(f)$ is set to the complexity of $C$ for $n_C$ and $b_C$,
to $0$ for $b_p$,  and to the sum of the complexities of $C_1$, $R$, $C$, and $C_2$, for $\textit{dist}(C_1,R{:}C,C_2)$.
Only features with cost bounded by $k$ are allowed in $\F^k$.

\section{Computational Model: Summary}

The steps for computing general plans are then:
\begin{enumerate}[1.]
  \item a sample set $\S$ is computed from a few instances $P$ of a given
    generalized problem $\Q$,
  \item a pool of features $\F$ is obtained from the predicates in the instances, %used in the instances $P$,
    the grammar, a  bound $k$, and the sample $\S$ (used for
    pruning),
  \item an assignment $\sigma$ of $T(\S,\F)$ or $T_G(\S,\F)$
    that minimizes $\sum_{f \in F_{\sigma}} cost(f)$ is obtained using a
    Max SAT solver,
  \item features $F$ and abstract actions $A_F$ are extracted from $\sigma$,
    %as $F_{\sigma}$ and $A_{\sigma}$,
  \item the abstraction $Q_F=\tup{V_F,I_F,G_F,A_F}$ is defined with initial and goal
    conditions $I_F$ and $G_F$ provided by hand to match $\Q$,
  \item the FOND problem $Q^+_F$, constructed automatically from $Q_F$, is solved with an off-the-shelf
    FOND planner.
\end{enumerate}

The policy $\pi$ that results from the last step deals with propositional symbols that correspond to atoms $p$ and $n=0$ for boolean
and numerical features $p$ and $n$ in $F$. When this policy is applied to an instance $P$ of $\Q$, the  value of the features $f$
is obtained from the functions $\phi_f(s)$ that they denote. For example, if $n=n_C$ for a concept $C$, then $\phi_{n}(s)=|C^s|$.

If we write $A \vDash B$ to express that all the states that satisfy $A$, satisfy $B$,
the formal guarantees can be expressed as:

\begin{theorem}
If the abstract actions $A_F$ that are sound relative to the sample set $\S$  are  sound,
then a   policy $\pi$ that solves  $Q^+_F$  is guaranteed to solve  all instances $P$ of $\Q$
with initial and goal situations $I$ and $G$ such that $I \vDash  I_F$ and $G_F \vDash G$.
\end{theorem}

% This means that a policy $\pi$ obtained with a FOND planner from $Q^+_F$
% solves the generalized problem $\Q$, provided that the abstract actions
% learned are sound relative to $\Q$ and that $\Q$ is restricted to the instances $P$ whose
% initial and goal conditions comply with   $I_F$ and $G_F$.

\Omit{
For the policy $\pi$  to be correct, however, the abstract actions do not have to be sound relative
to \emph{all} reachable states $s$ in the instances $P$ of $\Q$, it  suffices if they are sound on the
states $s$ that are reachable with the policy. We see an example of this below.
}

% \textcolor{red}{\bf ** NEXT IS UNCLEAR **}
% ok, I removed this. It's not strictly needed and we don't use it
\Omit{
Since we do not use a language for specifying $\Q$ formally, we will
take the formulas $I_F$ and $G_F$  as a partial formal specification of
$\Q$. The theorem then says that if soundness over the samples, ensure
formal soundness over $\Q$, the policy that solves the FOND problem $Q^+_F$,
solves $\Q$. The result holds whether $T$ or $T_G$ (goal marking) is used in step 3.
In principle, we could learn the formulas $I_F$ and $G_F$ from the sample as well,  but with different guarantees.
}

\section{Experimental Results}

We evaluate the computational model on four generalized problems $\Q$.
For each $\Q$, we select a few  ``training''  instances $P$ in $\Q$ by hand,
from which   the sample sets $\cal S$ are drawn.  $\S$ is constructed by collecting  the first $m$ states
generated by  a breadth-first search, along with  the states generated in an \emph{optimal} plan.
The plans ensure that $\cal S$ contains some goal states and provide the state transitions that are marked
as goal relevant when constructing  the theory $T_G({\cal S,\cal F})$, which is the one used in the experiments.
$\S$ is closed by fully expanding the states selected.  The value of $m$ is chosen so that
the resulting number of transitions in $\S$, which depends on the branching factor,
is around 500. The  bound $k$ for $\F=\F^k$ is set to $8$.
Distance features $dist$  are used only in the last problem. The Weighted-Max Solver is Open-WBO \cite{martins2014open}
and the FOND planner is SAT-FOND \cite{tomas:fond-sat}. The translation from $Q'_F$ to $Q^+_F$ is very fast, in  the order
of $0.01$ seconds in all cases.  The whole computational pipeline summarized by the steps 1--6 above is processed on Intel Xeon E5-2660 CPUs
with time and memory cutoffs of 1h and 32GB. Table~\ref{tab:exp-result-data} summarizes the relevant data for the problems,
including the size of the CNF encodings corresponding to the theories $T$ and $T_G$.

\begin{table*}[t]
  \centering
  %\resizebox{.95\textwidth}{!}{
    \begin{tabular}{lrrrrr@{}rrrrrrrr}
      \toprule
                     &     &            &            & \multicolumn{2}{c}{$T(\S,\F)$} && \multicolumn{2}{c}{$T_G(\S,\F)$} \\
      \cmidrule{5-6}
      \cmidrule{8-9}
                     & $n$ & $\abs{\S}$ & $\abs{\F}$ &   $np$ &   $nc$ &&   $np$ &   $nc$ & $t_{\text{SAT}}$ & $\abs{F}$ & $\abs{A_F}$ & $t_{\text{FOND}}$ & $\abs{\pi}$ \\
      \midrule
      $\Q_{clear}$   &   1 &        927 &        322 &   535K &  59.6M &&   7.7K &   767K &             0.01 &         3 &           2 &              0.46 &           5 \\
      $\Q_{on}$      &   3 &        420 &        657 &   128K &  25.8M &&  18.3K &   3.3M &             0.02 &         5 &           7 &              7.56 &          12 \\
      $\Q_{gripper}$ &   2 &        403 &        130 &    93K &   4.7M &&   8.1K &   358K &             0.01 &         4 &           5 &            171.92 &          14 \\
      $\Q_{reward}$  &   2 &        568 &        280 &   184K &  11.9M &&  15.9K &   1.2M &             0.01 &         2 &           2 &              1.36 &           7 \\
      \bottomrule
    \end{tabular}
  %}
  \caption{Results: $n$ is number of training instances $P$,
    $\abs{\S}$ is number of transitions in $\S$,
    %$k$ is max.\ feature complexity,
    $\abs{\F}$ is size of feature pool,
    $np$ and $nc$ are  numbers of propositions and clauses in $T(\S,\F)$ and $T_G(\S,\F)$,
    $t_{\text{SAT}}$ is time for SAT solver on $T_G$,
    $\abs{F}$ and  $\abs{A_F}$ are  number of
    selected features and abstract actions,
    $t_{\text{FOND}}$ is time for planner, and
    $\abs{\pi}$ is  size of the resulting policy.
    %$time_{\text{SAT}}$, $T_{\text{FOND}}$: computation time (sec.) of the SAT and FOND solvers, resp.;
    %$T_{\text{trans}}$: time (sec.) to generate the FOND problem $Q^+_F$;
    %$\#Act$, $\#At$: number if actions and atoms in $Q^+_F$;
    Times in seconds. %figures that depend on the SAT theory are given for $T_G(\S,\F)$.
  }
  \label{tab:exp-result-data}
\end{table*}

% For each problem: 1--6 details that vary from problem to problem. Follow same structure:
% $\Q$, $P$s, primitive predicates, number of atoms and clauses  of $T$ and/or $T^G$,
% SAT times,  $F$, $|A_F|$;  number of FOND atoms and actions in $Q^+_F$ (include those in $F$, $A_F$ plus translation), $I_F$ and $G_F$,
% FOND time, resulting policy.  Numbers can go all in a table.

\paragraph{Clearing a block.}
$\Q_{clear}$ contains the Blocksworld instances  with goals of the form  $clear(x)$
and stack/unstack actions. The primitive predicates, i.e., those appearing in the  instances $P$ of $\Q_{clear}$,
are $on(\cdot,\cdot)$, $clear(\cdot)$, $ontable(\cdot)$, $holding(\cdot)$, and $handempty$.
For this problem, a  single training instance $P$  with  5 blocks suffices to learn an abstract
model from  which a general plan is computed. The  set of features $F$ learned from the theory $T_G$ is:
\begin{enumerate}[--]
  \item $H: holding$ (whether some block is being held),
  \item $X: holding \sqcap C_x$ (whether block $x$ is being held),
  \item $n(x): \abs{\exists\,on^+ . C_x}$, (number of blocks above block $x$).
\end{enumerate}

\noindent The set $A_F$  of abstract actions learned is:\footnote{Feature and action
  names are provided to make their meaning explicit; the meaning follows from their syntactic form.}

\begin{enumerate}[--]
  \item $\text{put-aside} = \abst{\neg X, H}{\neg H}$,
  \item $\text{pick-above-$x$} = \abst{\neg X, \neg H, n(x) > 0}{H, n(x)\mminus}$.
\end{enumerate}

The  abstraction $Q_F=\tup{V_F,I_F,G_F,A_F}$ for $\Q_{clear}$ is set with
$I_F = \{\neg H, \neg X, n(x) > 0\}$ and $G_F=\{n(x)=0\}$.
As mentioned above, since no action increments the variable $n(x)$,
the strong-cyclic solutions of $Q'_F$ solve $\Q_{clear}$.
One such policy is found with the FOND planner in 0.46 seconds.
The plan implements a loop that applies the pick-above-$x$
action followed by put-aside, until $x$ becomes clear.

% Action 1: PLACE-ASIDE
%         PRE: NOT holding(a), bool[holding]
%         EFFS: NOT bool[holding]
%
% Action 2: PICK-FROM-A
%         PRE: NOT bool[holding], NOT holding(a), n(a) > 0
%         EFFS: DEC n(a), bool[holding]

% Loop:
%   Action 2 [Pick block above x]
%   If n(a) = 0 Break
%   Action 1 [Put block aside]
%

\paragraph{Stacking two blocks.}
$\Q_{on}$ consists  of Blocksworld instances with goals of the form $on(x,y)$,
and initial situations  where the blocks $x$ and $y$ are in \emph{different towers}.
The primitive predicates are the same as in $\Q_{clear}$.
Three training instances are used to learn $F$ from $T_G$;
$F$ contains
%
% Using the above Blocksworld encoding and only 3 different training
% instances with up to 14 blocks, an abstract model involving 3 boolean
% features, 2 numerical features, and 7 actions is learnt.
the boolean features $E$, $X$ and $G$, for the gripper being empty, the
block $x$ being held, and $x$ being on block $y$, and the
numerical features $n(x)$ and $n(y)$ that count the number of blocks
above $x$ and $y$ respectively.
% The DL  definition of the  features $X$, $n(x)$, and $n(y)$ are as above
% while $E$ is $handempty$ and $G$ stands for $(\exists\,on.C_y)  \sqcap C_x$.
% Interestingly, this last feature is the concept-based representation
% of the goal $on(x,y)$.
%
%% blocks
%% 5 bool[handempty] 0 n(a) 1 n(b) 1 holding(a) 0 on(a,b)_2 0
%% 5 bool[handempty] 1 holding(a) 0 n(a) 1 n(b) 1 on(a,b)_2 0
%% 4 holding(a) 0 n(a) 0 n(b) 1 on(a,b)_2 1
%% 7
%% action_1 ; put held block (diff. from a and b) aside
%% 4 bool[handempty] 0 holding(a) 0 n(a) 0 on(a,b)_2 0
%% 1 bool[handempty] 1
%% action_2 ; put aside
%% 5 bool[handempty] 0 holding(a) 0 n(a) 1 n(b) 1 on(a,b)_2 0
%% 1 bool[handempty] 1
%% action_3 ; pick block a
%% 5 bool[handempty] 1 holding(a) 0 n(a) 0 n(b) 0 on(a,b)_2 0
%% 2 bool[handempty] 0 holding(a) 1
%% action_4 ; put a (being held) aside
%% 5 bool[handempty] 0 holding(a) 1 n(a) 0 n(b) 1 on(a,b)_2 0
%% 2 bool[handempty] 1 holding(a) 0
%% action_5 ; pick block different from b that is above a
%% 5 bool[handempty] 1 holding(a) 0 n(a) 1 n(b) 1 on(a,b)_2 0
%% 2 bool[handempty] 0 n(a) 0
%% action_6 ; pick block different from a that is above b
%% 5 bool[handempty] 1 holding(a) 0 n(a) 0 n(b) 1 on(a,b)_2 0
%% 2 bool[handempty] 0 n(b) 0
%% action_7 ; put a directly on b
%% 5 bool[handempty] 0 holding(a) 1 n(a) 0 n(b) 0 on(a,b)_2 0
%% 4 bool[handempty] 1 holding(a) 0 n(b) 1 on(a,b)_2 1
The learned set of actions $A_F$ is:
\begin{enumerate}[--]
  \item $\text{pick-ab-$x$}\!=\!\abst{E,\neg X,\neg G,n(x)\!\!>\!0,n(y)\!>\!0}{\neg E,n(x)\mminus}$,
  \item $\text{pick-ab-$y$}\!=\!\abst{E,\neg X,\neg G,n(x)\!=\!0,n(y)\!\!>\!0}{\neg E,n(y)\mminus}$,
  \item $\text{put-aside-1}=\abst{\neg E,\neg X,\neg G,n(x)=0}{E}$,
  \item $\text{put-aside-2}=\abst{\neg E,\neg X,\neg G,n(x)>0,n(y)>0}{E}$,
  \item $\text{pick-$x$}=\abst{E,\neg X,\neg G,n(x)=0,n(y)=0}{\neg E,X}$,
  \item $\text{put-$x$-aside}\!=\!\abst{\neg E,X,\neg G,n(x)=0,n(y)>0}{E,\neg X}$,
  \item $\text{put-$x$-on-$y$} = \langle \neg E, X, \neg G, n(x)=0, n(y)=0; E, \neg X,$ $G, n(y)\pplus \rangle$.
\end{enumerate}

The abstraction $Q_F$  with $I_F=\{E,\neg X,\neg G,n(x)>0,$ $n(y)\!>\!0\}$
% and $G_F = \{ \neg X, G, n(x)=0, n(y)\!>\!0\}$,
and $G_F = \{G\}$ is translated into the FOND problem $Q_F^+$
which is then solved by the planner in 7.56 seconds.
The resulting policy solves the generalized problem $\Q_{on}$:
it implements a loop to clear block $x$, followed by a loop
to clear block $y$, picking then $x$ and placing it on $y$.

\paragraph{Gripper.}
$\Q_{gripper}$ involves  a robot with grippers whose goal
is to move a number of balls from one room into a target
room $x$. Each gripper may carry one ball at a time.
The STRIPS predicates are $at\text{-}robby(l)$, $at\text{-}ball(b,l)$,
$free(g)$, $carry(b,g)$ that denote, respectively, whether the robot
(the ball) is at location $l$, whether gripper $g$ is free, and whether
gripper $g$ carries ball $b$, plus unary type predicates $room$, $ball$,
and $gripper$. % The type predicates are also included.

Pipeline 1--6 is fed with two instances $P$ from $\Q_{gripper}$  with two
rooms each, one with 4 balls and the other with 5. The set of features that
is learned  from the theory $T_G$ is:
\begin{enumerate}[--]
  \item $X: at\_robby \sqcap C_x$ (whether robby is in target room),
  \item $B: |\exists\,at . \neg C_x|$ (number of balls not in target room),
  \item $C: |\exists\,carry . C_u|$ (number of balls carried),
  \item $G: \abs{free}$ (number of empty grippers).
\end{enumerate}

\noindent The set of  abstract actions $A_F$  that is learned is:

\begin{enumerate}[--]
  \item $\text{drop-ball-at-$x$}=\abst{C>0, X}{C\mminus, G \pplus}$,
  \item $\text{move-to-$x$-half-loaded}\! =\!  \abst{\neg X, B=0, C>0, G\!>\!0}{X}$,
  \item $\text{move-to-$x$-fully-loaded} = \abst{\neg X, C>0, G=0}{X}$,
  \item $\text{pick-ball-not-in-$x$} = \abst{\neg X, B > 0, G > 0}{B\mminus, G\mminus, C\pplus}$,
  \item $\text{leave-$x$} = \abst{X, C=0, G > 0}{\neg X}$.
\end{enumerate}

% Once again, although this set of abstract actions  is not complete,
% it is sufficient for generating a general plan.
The abstraction $Q_F$ with $I_F\!=\!\{ B\!>\!0, \neg X, G\!>\!0, C\!=\!0\}$ and $G_F\!=\!\{ B\!=\!0, C\!=\!0 \}$ is
translated into $Q^+_F$ which is solved by the planner in 171.92 seconds.
The resulting policy provably solves $\Q_{gripper}$, meaning that
it works for any  number of grippers and balls.
\Omit{ %% Add if space
The policy picks the balls in the source room, one by one,
until the grippers are full or there are no more balls.
It then moves to the target room, with one of the two move actions above,
according to whether the grippers are fully loaded or not,
and then drops the balls one by one, until the grippers
are all empty.  It then moves back to the other room if there
are still  balls to be moved, repeating the process.
}

% Like before, the extra conditions
% that are not strictly required (in the move actions), do not preclude
% the solution of the problem, that results in a general plan that
% works  for robots with any  number of grippers and  any number of balls.

\paragraph{Collecting rewards.}
$\Q_{reward}$ contains instances where  an agent needs to navigate a rectangular
grid to pick up  rewards spread on the grid  while
avoiding \emph{blocked} cells. This is a variation of an
example  by \citeay{garnelo2016towards}.
The STRIPS instances have the primitive predicates $reward(\cdot)$, $at(\cdot)$, $blocked(\cdot)$
and $adjacent(\cdot,\cdot)$ that denote the position of the rewards, of
the agent, of the blocked cells, and the grid topology respectively.
Pipeline 1--6  is fed with two instances of the problem of sizes $4 \times 4$
and $5\times 5$, and having different distributions of blocked cells and rewards.
Two numerical features are learned:
\begin{enumerate}[--]
  \item $R: \abs{reward}$ (number of remaining rewards),
  \item $D:\! dist(at, adjacent{:}\neg blocked, reward)$ (distance from current
    cell to closest cell with reward, traversing adjacent, unblocked cells only).
\end{enumerate}
The learned set of abstract actions is:
\begin{enumerate}[--]
  \item $\text{collect-reward} = \abst{D=0, R > 0}{R\mminus, D\pplus}$,
  \item $\text{move-to-closest-reward} = \abst{R>0, D>0}{D\mminus}$.
\end{enumerate}

The resulting  abstraction $Q_F$ with $I_F = \{ R>0, D>0 \}$ and $G_F=\{R=0\}$
is translated into $Q^+_F$ which is solved by the planner in 1.36 seconds.
The policy moves the agent one step at a time towards the uncollected reward
that is closest, as measured by the numerical feature $D$. Once the reward is
reached, the reward is consumed, and the process repeats until there are no
more rewards to be collected.

\section{Discussion}

The computational  bottleneck of the approach
is  in the size of the SAT  theories used to derive the features
and the actions.
% This puts a limit on the complexity of the
% feature grammars, the bound, and the number of state transitions in the sample,
% as they  result otherwise in SAT theories that are just too large (see Table~\ref{tab:exp-result-data}).
This is the   reason why we have chosen to use the more compact theories $T_G$
in the experiments.  Additional ideas, however, are required
to improve scalability and to make the computational model
captured by  steps 1--6 more robust.
% One possibility that we have not explored
% is the use of incomplete (suboptimal) Max SAT solvers.

From the point of view of expressivity, it is not clear how restrictive
is the assumption  that the  features can be obtained from a pool of  features
defined from the primitive predicates and a general grammar.
%A second limitation is the assumption that
%the features can be obtained from a large pool of candidate  features defined
%from the primitive predicates and a  general grammar.
%It is not fully clear whether this assumption is reasonable in general.
A relevant argument made by \citeay{bonet:ijcai2018} is that generalized
problems over domains with bounded width \cite{nir:ecai2012} have compact
policies in terms of features $f$ whose value $\phi_f(s)$ can be computed
in polynomial time for any state $s$. It is an open question whether
such features are  captured by the proposed grammar, or a  suitable variation.
The feature language, however, seems adequate for  dealing with arbitrary goals,
once  \emph{goal predicates} are added to the set of primitive predicates.
%Goal predicates are copies $p_G$ of the primitive predicates $p$ that appear
%in the goal and which are evaluated in the goal  \cite{martin-geffner:generalized}.
Goal predicates are ``copies'' $p_G$ of the primitive predicates $p$ that
appear in the goal, and have fixed interpretation\footnote{If $p$ is
  a primitive predicate that appears in the goal, for any state $s$ and
  tuple $\bar u$ of objects, $s\vDash p_G(\bar u)$ iff $p(\bar u)$
  holds in the goal.}
\cite{martin-geffner:generalized}.

Finally, our formulation can be extended to handle non-deterministic actions.
For this, it is sufficient to replace the formula \eqref{eq:bridge1}
that links $D_1$ and $D_2$ atoms  with a formula asserting that,
if two states $s$ and $t$ are indistinguishable (i.e., $\neg D_1(s,t)$), then
for each action $a$ in $s$, there is an action $b$ in  $t$
such that the set of transitions $(t,b,t')$ generated by $b$ in $t$
cannot be distinguished from the set of transitions $(s,a,s')$ generated by $a$ in $s$.
For this, transitions $(s,s')$ in the sample would need to be tagged with the actions
that generated them as  $(s,a,s')$.

\section{Summary and Future Work}

We have introduced a scheme for computing general plans that mixes
learning and planning: a \emph{learner} infers an abstraction made
up of features and abstract actions by enforcing %a notion of
soundness and completeness over the samples, and a \emph{planner}
uses the abstraction, suitably transformed, to compute general plans.
The number of samples required for obtaining correct general plans is
small, as the learner does not have to produce plans; it just
has to learn the relevant features for the planner to track.
Unlike purely learning approaches, the features and the policies
are transparent, and the scope and correctness
of the resulting general plans can be assessed.

There is  an interesting  relation between sound and complete abstractions, on the one hand,
and the ideas of dimensionality reduction and embeddings in machine learning, on the other
\cite{hamilton:embeddings}. Sound and complete abstractions  map  the  states of the problem
instances,  whose size is not bounded, into valuations over a small and bounded set of features,
while preserving the   essential properties of states; namely,  how they are changed by actions
and whether they denote  goal states or not.
% The notion of \emph{sound} and \emph{complete} abstract action sets expresses this requirement
% that is enforced by  the SAT formulation.
An interesting challenge for the future  is to generalize the proposed methods
to other contexts, such as learning from ``screen states'' in video games,
where the sampled states have no known structure and there are no primitive
predicates. One way for achieving this would be precisely by learning embeddings
that yield sound and complete abstractions or suitable approximations.

%Finally, problems involving non-deterministic transitions may be
%accounted for with a simple change to the SAT theory.
%It would be sufficient to replace the formula \eqref{eq:bridge1}
%that links $D1$ and $D2$ with a formula that establishes that when
%$s$ and $t$ are indistinguishable (i.e.\ $\neg D1(s,t)$), then
%for each action $a$ for $s$, there is an action $b$ for $t$ whose
%set of transitions on $t$ is indistinguishable from the set
%of transitions of $a$ on $s$; namely, for each transition $(s,a,s')$
%there is a transition $(t,b,t')$ with $\Delta_f(s,s')=\Delta_f(t,t')$
%for each feature $f$, and vice versa.
%To do this, transitions $(s,s')$ in the sample would need to be
%tagged with action labels as $(s,a,s')$.

%\bigskip
\subsection*{Acknowledgments}
B.\ Bonet is partially funded by 2018 C\'atedra de Excelencia UC3M-Banco Santander award.
G.\ Franc\`{e}s is funded by the SSX project, European Research Council.
H.\ Geffner is partially funded by grant TIN-2015-67959-P, MINECO, Spain.

\bibliographystyle{aaai}
\bibliography{control}
\end{document}